\pdfoutput=1

\documentclass[11pt]{article}

\usepackage{acl}

\usepackage{latexsym, graphicx, float, amsmath, tabularx, subcaption}
\usepackage{booktabs} 
\usepackage{threeparttable} 
\usepackage{times}
\usepackage{latexsym}
\usepackage{hyperref}
\usepackage{url}
\usepackage{multirow}
\usepackage[T1]{fontenc}

\usepackage[utf8]{inputenc}

\usepackage{microtype}

\usepackage{inconsolata}
\usepackage{soul}
\usepackage{subcaption}
\newcommand{\ourmethod}{{\fontfamily{qag}\selectfont{\small{EvolvTrip}}}}
\newcommand{\ourmethodtitle}{{\fontfamily{qag}\selectfont{EvolvTrip}}}
\newcommand{\ourdataset}{{\fontfamily{qag}\selectfont{\small{LitCharToM}}}}
\newcommand{\ourdatasettitle}{{\fontfamily{qag}\selectfont{LitCharToM}}}

\definecolor{tablewhite}{HTML}{EEEEEE}

\title{\ourmethodtitle: Enhancing Literary Character Understanding with Temporal Theory-of-Mind Graphs}

\newcommand*\samethanks[1][\value{footnote}]{\footnotemark[#1]}

\author{ 
Bohao Yang\textsuperscript{1}\thanks{\quad \small Equal contribution.},
Hainiu Xu\textsuperscript{2}\samethanks\space,
Jinhua Du\textsuperscript{3}\space,
Ze Li\textsuperscript{4}\space,
Yulan He\textsuperscript{2,5}\thanks{\quad \small Corresponding authors}\space,
\textbf{Chenghua Lin\textsuperscript{1}}\samethanks \space\; \\
\textsuperscript{1} The University of Manchester\vspace{-0.5mm} 
\textsuperscript{2} King’s College London\vspace{-0.5mm}\\
\textsuperscript{3} Huawei London Research Centre\vspace{-0.5mm}
\textsuperscript{4}Huawei Technologies Co., Ltd.\vspace{-0.5mm}
\textsuperscript{5} The Alan Turing Institute\vspace{-0.5mm}
\\
\texttt{
bohao.yang-2@postgrad.manchester.ac.uk
}\vspace{-0.1mm}
\texttt{
chenghua.lin@manchester.ac.uk,
}\vspace{-0.5mm}\\
\texttt{
\{jinhua.du, lize23\}@huawei.com
}\vspace{-0.5mm} 
\texttt{
\{hainiu.xu, yulan.he\}@kcl.ac.uk
}\vspace{-0.5mm} \\
}

\begin{document}
\maketitle
\begin{abstract}
A compelling portrayal of characters is essential to the success of narrative writing. For readers, appreciating a character’s traits requires the ability to infer their evolving beliefs, desires, and intentions over the course of a complex storyline, a cognitive skill known as Theory-of-Mind (ToM).
Performing ToM reasoning in prolonged narratives requires readers to integrate historical context with current narrative information, a task at which humans excel but Large Language Models (LLMs) often struggle.
To systematically evaluate LLMs' ToM reasoning capability in long narratives, we construct~\ourdataset, a benchmark of character-centric questions across four ToM dimensions from classic literature. Further, we introduce~\ourmethod, a perspective-aware temporal knowledge graph that tracks psychological development throughout narratives. Our experiments demonstrate that~\ourmethod~consistently enhances performance of LLMs across varying scales, even in challenging extended-context scenarios.~\ourmethod~proves to be particularly valuable for smaller models, partially bridging the performance gap with larger LLMs and showing great compatibility with lengthy narratives. Our findings highlight the importance of explicit representation of temporal character mental states in narrative comprehension and offer a foundation for more sophisticated character understanding.
Our data and code are publicly available at
\url{https://github.com/Bernard-Yang/EvolvTrip}.

\end{abstract}

\begin{figure}[!t]
\centering
\includegraphics[width=0.8\columnwidth]{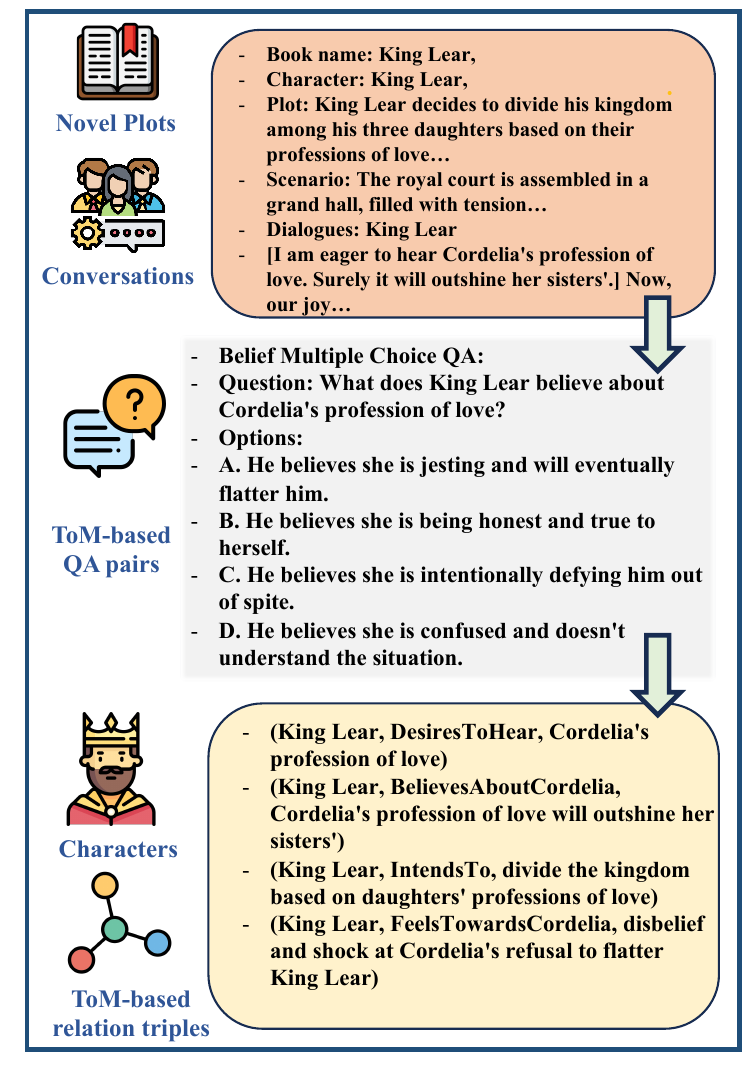}
\caption{Our ToM-based character understanding pipeline, showing how novel plots and character conversations are transformed into multiple-choice questions and structured relation triples that represent character mental states across belief, desire, intention, and emotion dimensions.
}
\vspace{-1em}
\label{fig:intro}
\end{figure}
\section{Introduction}

Theory of Mind (ToM), the capability to infer others' mental states such as beliefs, desires, and intentions, is substantial for narrative comprehension \cite{premack1978does, apperly2010mindreaders}, where understanding charaters' motivations and predicting their behaviors across extended storylines demands readers to construct rich mental models of each character.
Specifically, ToM reasoning over prolonged narratives requires comprehensive contextualization of accumulated knowledge about characters' backgrounds, personalities, and past experiences with their current circumstances~\cite{davis1983measuring, harwood2006conflicting, apperly2010mindreaders}. When engaging with narratives, humans constantly construct and update models of characters' mental states throughout the storyline, allowing for tracking psychological development and drawing connections between past experiences and present behaviors \cite{schneider2001toward}. Such a temporal and evolutionary dimension of understanding, which is crucial for deep character comprehension, remains underexplored in computational approaches.
Despite the increasing sophistication of Large Language Models (LLMs), research reveals significant limitations in their ToM reasoning capabilities, particularly in complex narrative contexts \cite{nematzadeh-etal-2018-evaluating, gandhi2023understanding,DBLP:conf/lrec/TraceyRC0DDDGMM22, Ullman2023LargeLM, zhou2025essence}. 

Perspective-taking, which involves inferring what different characters perceive and know based on their unique vantage points, constitutes a critical aspect of human ToM reasoning \cite{davis1983measuring, harwood2006conflicting}. For readers of novels, perspective-taking is enriched by accumulated knowledge of characters' backgrounds and past experiences. However, existing computational approaches to ToM reasoning often neglect this crucial dimension, instead focusing on isolated scenarios without sufficient global context~\cite{wilf2023think, huang2024notion,hou2024timetom, jung-etal-2024-perceptions, zhou2025essence}.
Prior ToM benchmarks like CharToM~\cite{zhou2025essence} evaluate understanding through brief vignettes with limited character history. 

In light of the need for a benchmark that examines LLMs' long-context ToM reasoning capabilities, we construct \ourdataset. \ourdataset~is built upon classic literary narratives with characters that possess rich experiences developed over time through multiple interactions and evolving circumstances. This temporal dimension allows us to evaluate models' ability to keep track of characters' psychological evolutions, an essential capability for human-like narrative comprehension.

To enhance LLMs' ToM reasoning capabilities in long narratives, we propose \ourmethod~a novel framework for understanding fictional characters via temporal-aware structured mental state representation. 
While previous works such as  PerceptToM and EnigmaToM~\cite{jung-etal-2024-perceptions, xu2025enigmatomimprovellmstheoryofmind} focus on visual perception, \ourmethod~models complex mental states informed by characters' backgrounds, histories, and accumulated experiences. By encoding these perspective-aware mental states as structured triples within a temporal knowledge graph, \ourmethod~enable LLMs to reason about character psychology with contextual richness more closely resembling human ToM processes during narrative comprehension.
Empirical results show that \ourmethod~brings significant performance improvements in long-context ToM reasoning to a range of LLMs. \ourmethod~is particularly effective in modeling ToM in extended-context scenarios with corss-plot narrative contents. Further, \ourmethod~is also effective when used with smaller LLMs, partially bridging the performance gap with larger architectures and demonstrating enhanced resilience when processing longer narratives.


Our contributions can be summarised as follows:
\begin{itemize}
\item We construct \ourdataset, a character-centric benchmark for evaluating ToM reasoning in literary contexts using classic novels. \ourdataset~provides rich scenarios with complex social dynamics and long-term narrative dependencies, enabling comprehensive assessment of contextual understanding.

\item We introduce a perspective-aware temporal knowledge graph with entity-guided character linking. Our knowledge graph represents characters' mental states as structured triples tagged with temporal markers and connects character instances across narrative segments. 

\item We propose \ourmethod, a neuro-symbolic approach for enhancing ToM reasoning. \ourmethod incorporates structured representation of characters' evolving mental states, which significantly improves LLMs' performance on character-centric ToM reasoning that require deep contextual understanding. 
\end{itemize}

\begin{figure*}[htb]
\centering 
\includegraphics[width=0.9\linewidth]{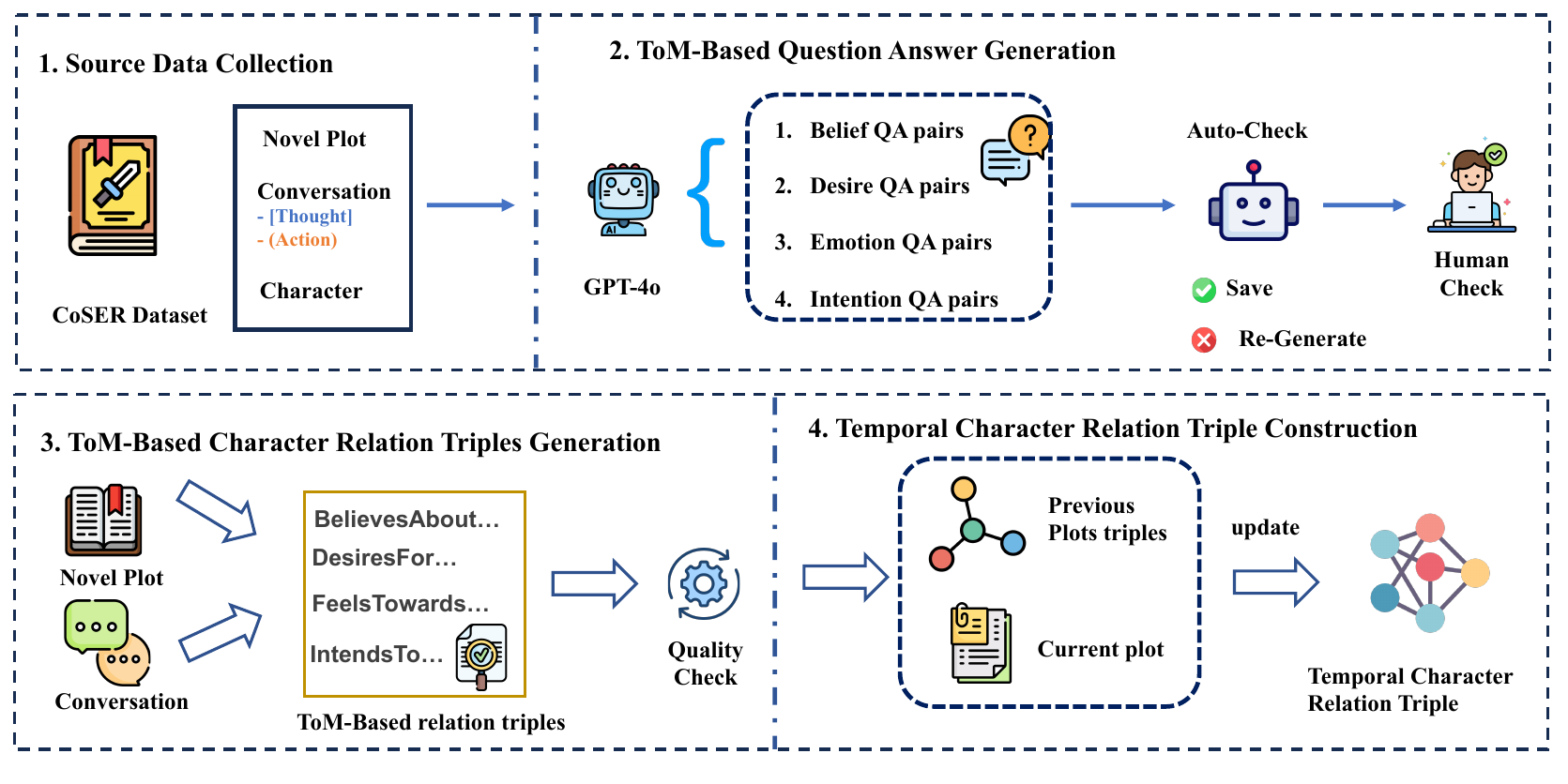}
\caption{Our ToM-based character understanding pipeline: (1) Source data collection from CoSER Dataset including novel plots and character conversations with [Thought] and (Action) annotations, (2) GPT-4o generation of belief, desire, emotion, and intention QA pairs with two-stage verification, (3) Extraction of BelievesAbout, DesiresFor, FeelsTowards, and IntendsTo relation triples, and (4) Temporal knowledge graph construction by integrating previous and current plot information.
    }
\vspace{-0.5em}
\label{fig:model}
\end{figure*}

\section{Related Work}

\subsection{Theory of Mind Evaluation in LLMs}

Numerous benchmarks have been developed to evaluate ToM capabilities in LLMs by simulating psychological and cognitive experimental designs. Early benchmarks like ToMi \cite{DBLP:conf/emnlp/NematzadehBGGG18} focused on evaluating models' ability to reason about basic beliefs. This foundation was extended by SocialIQA \cite{DBLP:conf/emnlp/SapRCBC19}, which specifically tests social and emotional intelligence. More advanced ToM reasoning has been explored in Hi-ToM \cite{DBLP:conf/emnlp/WuHJM0D23}, which assesses higher-order recursive reasoning about others' beliefs.
Recent benchmarks have diversified the evaluation contexts, with FANToM \cite{DBLP:conf/emnlp/0002SZ0K0S23} stress-testing ToM within conversational settings and OpenToM \cite{DBLP:conf/acl/XuZZD024} incorporating explicit personality traits and preferences. Comprehensive evaluation platforms like ToMBench \cite{DBLP:conf/acl/0002WZWBJCHLXH24} encompass multiple tasks that target 31 distinct social cognitive abilities.
Despite their wide coverage, these benchmarks share common limitations. Most rely heavily on pre-determined rules and templates for scenario generation \cite{DBLP:conf/emnlp/NematzadehBGGG18,DBLP:conf/emnlp/LeBN19}, which can introduce predictable patterns and spurious correlations, potentially leading to the Clever Hans phenomenon \cite{DBLP:journals/corr/abs-1902-10178}. Moreover, they typically feature brief, isolated scenarios that fail to capture the complexity of social relationships and interactions that characterize real-world ToM reasoning, overlooking the importance of comprehensive contextual understanding that spans extended narrative timeframes.

\noindent\textbf{Character Understanding in Narrative Comprehension}
There has been consistent efforts in character-centric narrative understanding, with works like NarrativeQA \cite{kovcisky2018narrativeqa}, LitBank \cite{bamman-etal-2019-annotated, sims-etal-2019-literary, bamman-etal-2020-annotated}, LiSCU \cite{brahman-etal-2021-characters-tell}, and PeQA \cite{xu-etal-2022-fantastic} developing question-answering frameworks for longer narrative contexts. These approaches primarily evaluate surface-level comprehension rather than deeper understanding of characters' mental states and psychological development.
The psychology literature consistently shows that human readers construct rich mental models of fictional characters' beliefs and intentions \cite{apperly2010mindreaders}, tracking these mental states across extended narratives. This cognitive process relies heavily on accumulated knowledge of characters' backgrounds, histories, and evolving psychological states—aspects that most computational approaches have not adequately modeled.

\noindent\textbf{Knowledge Representation for ToM Reasoning}
Knowledge bases for representing mental states and social reasoning have evolved from general-purpose semantic networks like ConceptNet \cite{liu2004conceptnet} to more specialized representations. Event2Mind \cite{rashkin2018event2mind} introduced event-based knowledge graphs that capture characters' intentions and reactions, while ATOMIC \cite{sap2019atomic} models if-then relationships for simple social events.
Recent approaches include entity state tracking in procedural contexts \cite{tandon2020dataset, zhang2023openpi2}, though these have not been specifically applied to character understanding in extended narratives. In the mean time, Neural knowledge bases like COMET is developed \cite{bosselut2019comet}, which generate commonsense inferences about social situations, but lack the temporal depth needed for character tracking across narrative arcs.





\section{Dynamic Character Understanding through Evolving Mental State Triplets}
\label{sec:dataset_construction}

We introduce the construction of the \ourdataset~benchmark and the design of \ourmethod~framework for evaluating Theory-of-Mind comprehension in literary narratives. \ourmethod~(Evolving Triplets) is a structured knowledge representation approach that captures the dynamic evolution of character mental states across narrative arcs. Following the pipeline illustrated in Figure \ref{fig:model}, our construction methodology encompasses four integrated phases: (1) source data collection, (2) ToM-based question generation, (3) character relation triple extraction, and (4) temporal knowledge graph construction.

\subsection{\ourdatasettitle: Source Data Collection}
\ourdataset~builds upon the CoSER dataset\footnote{We use the Gutenberg branch of the CoSER dataset to ensure copyright compliance. \url{https://huggingface.co/datasets/Neph0s/CoSER-Books-Gutenberg}} \citep{wang2025coser}, which comprises 81 literary works from project Gutenberg. CoSER provides rich character-centric data including plot summaries, character profiles, and multi-dimensional dialogues. We further selected 20 books from CoSER that exhibit sophisticated character development, complex interpersonal dynamics, and narrative depth spanning multiple scenes. See Appendix~\ref{sec:dataset_statistics} for detailed statistics of \ourdataset.

We base our \ourdataset~on CoSER dataset because of its multi-dimensional representation of character dialogue, which includes \textit{verbal speech} (direct communications), \textit{actions} (physical behaviors denoted by parentheses), and \textit{thoughts} (internal cognitive processes denoted by brackets). This tripartite structure offers particular value for ToM analysis, as each dimension maps differently to mental state categories. \textit{Actions} reveal intentions and emotions (e.g., nods firmly suggests deliberate agreement). \textit{Thoughts} provide rich access to all four ToM dimensions, with strongest mapping to emotions (e.g., [I'm terrified]), followed by desires (e.g., [I wish I could leave]), intentions (e.g., [I'll confront him tomorrow]), and beliefs (e.g., [He's lying to everyone]). 
This structured representation enables \ourmethod~to extract both explicit and implicit mental states from complementary sources, where thoughts reveal deeper affective and cognitive layers, and actions reflect behavioral manifestations of internal states.

\subsection{\ourdatasettitle: ToM-Based Question Generation}
For each character participating in each plot's dialogues, we systematically generate ToM questions across four dimensions: belief, emotion, intention, and desire. We employ GPT-4o~\cite{OpenAI2024} to construct multiple-choice questions requiring reasoning about characters' mental states.

For each ToM dimension, GPT-4o examines multiple sources of information: the current plot content, conversation scenario, character dialogues (including the thoughts of current character), and summaries of previous plot segments. This comprehensive context allows the model to identify salient mental states across narrative progression, formulating complex questions with four answer options: one correct answer grounded in the character's depicted psychology and three plausible distractors representing common misinterpretations.
To ensure accuracy, we implement a two-stage verification process: initially, GPT-4o verifies all generated questions for logical consistency, clarity, and the presence of a single unambiguously correct answer. Subsequently, human annotators assess accuracy, difficulty level, and appropriateness. Notably, over 90\% of the entries are valid at the first generation attempt\footnote{See Appendix~\ref{app:quality} for detailed statistics on data quality control.}, demonstrating the effectiveness of our generation methodology. Questions identified as problematic during either verification stage undergo refinement or complete regeneration, followed by an additional verification process.
\subsection{\ourmethodtitle: Mental State Triple Extraction}
To provide a structured representation of characters' mental activities, \ourmethod~extracts character-centric mental state triples following a subject-predicate-object structure. The subject corresponds to the character, the predicate indicates the ToM dimension (e.g., BelievesAbout, FeelsTowards, IntendsTo, DesiresFor), and the object constitutes the content of the mental state.

For each narrative plot, we employ GPT-4o to generate triples by analyzing the multi-dimensional dialogue data through a perspective-taking lens, which distinguishes between information accessible to each character versus information they cannot know. This perspective-aware approach examines character thoughts that directly reveal mental states, character actions that imply underlying mental states, and verbal dialogues containing explicit statements about beliefs, emotions, intentions, or desires. By identifying events observable by a given character and excluding unobservable ones, this approach significantly alleviates the reasoning burden for LLMs, enabling more accurate mental state attribution. Predicates are specified to provide precise context, such as using BelievesAbout to indicate a belief concerning another entity or FeelsTowards to denote an emotion directed at someone.
For triple verification, GPT-4o conducts initial assessment of all generated triples for logical consistency with the narrative context, adherence to the correct triple format, and appropriate perspective constraints (ensuring characters only form mental states about information they could plausibly access). We then randomly select 40\% of triples for human expert verification, assessing their accuracy and relevance to the characters' depicted mental states. Triples identified as incorrect during either verification stage are regenerated and re-verified, ensuring high-quality knowledge representation. Detailed dataset quality statistics are provided in Appendix~\ref{app:quality}.


\subsection{\ourmethodtitle: Temporal Knowledge Graph Construction}
The core innovation of \ourmethod~is capturing the dynamic nature of character psychology throughout narratives. We construct a temporal knowledge graph where nodes represent characters or significant events, edges embody the generated triples with labels specifying the ToM dimension, and temporal tags associate each triple with specific plot numbers.
Each triple is tagged with the plot segment in which the mental state appears, enabling systematic tracking of psychological development. We establish inter-plot links between instances of the same character across different segments, facilitating analysis of how characters' mental states evolve in response to narrative developments.

To maintain psychological consistency, we provide GPT-4o the past mental states of each character when generating triples for new plot segments. This approach enables it to build upon established psychological profiles. For similar mental states concerning the same subject, \ourmethod~combines or refines them based on new information. When new information contradicts earlier states, we update the triples to reflect character development, clearly indicating the temporal transition to demonstrate how the character's perspective has evolved throughout the narrative.
This temporally linked representation provides a comprehensive view of character psychology that evolves organically through the narrative, capturing the dynamic nature of beliefs, emotions, intentions, and desires as they transform in response to story events.
\section{Experiments}

\subsection{Setup}
We conduct experiments on our multiple-choice Theory-of-Mind benchmark comprising 2,539 questions spanning four dimensions: belief, emotion, intention, and desire. All experiments use a standardized prompt template as detailed in Appendix~\ref{app:prompt}. To investigate models' ability to leverage contextual information for ToM comprehension, we vary the context lengths of story plots provided to the models, examining their performance with and without the structured triple representations generated by \ourmethod~.
For each question, models are evaluated in two settings: (1) standard prompting with only the narrative context and question, and (2) \ourmethod~-enhanced prompting where relevant mental state triples are included as additional context. This allows us to assess the impact of \ourmethod's explicit structured knowledge on models' ToM reasoning capabilities.


\vspace{0.5em}
\noindent\textbf{Evaluated LLMs.}~~We evaluate a diverse set of LLMs as our baselines, including GPT-4o and GPT-4o-mini \cite{openai2023gpt4}, accessed through official APIs. For the open-sourced LLMs, we include DeepSeek-R1 \cite{deepseekai2025deepseekr1incentivizingreasoningcapability}, Qwen2.5-72B-Instruct \cite{qwen2.5}, Llama3.3-72B-Instruct \cite{dubey2024llama}, DS-R1-Dist-Qwen-32B (DeepSeek-R1 distilled into a 32B Qwen architecture)~\cite{deepseekai2025deepseekr1incentivizingreasoningcapability}, Qwen3-32B~\cite{qwen3}, Qwen2.5-32B-Instruct~\cite{qwen2.5}, InternLM2.5-20B-Chat\cite{cai2024internlm2}, Qwen3-14B~\cite{qwen3}, Qwen2.5-14B~\cite{qwen2.5}, DS-R1-Dist-Qwen-14B~\cite{deepseekai2025deepseekr1incentivizingreasoningcapability}, Qwen3-8B~\cite{qwen3}, Qwen2.5-7B-Instruct~\cite{qwen2.5}, InternLM3-8B-Instruct~\cite{cai2024internlm2}, and InternLM2.5-7B-Chat~\cite{cai2024internlm2}. For each model, we test both a standard version and a triple-enhanced version (denoted as "w Triple") that incorporates structured mental state triples into the context. All models are accessed either through official APIs or using weights downloaded from Hugging Face repositories, in compliance with their terms of use.

\begin{table*}[htb]
    \footnotesize
    \centering
    \renewcommand{\arraystretch}{0.8}
    \begin{threeparttable}[b]
    \setlength{\tabcolsep}{15pt}  
    \begin{tabular}{lccccc}
     \toprule
    \multirow{2}{*}{\textbf{Models}} 
     & \textbf{Belief} & \textbf{Desire} & \textbf{Emotion} & \multicolumn{1}{c}{\textbf{Intention}} & \textbf{Avg} \\ 
    &\multicolumn{1}{c}{Acc.} & \multicolumn{1}{c}{Acc.} & \multicolumn{1}{c}{Acc.} & \multicolumn{1}{c}{Acc.} & \multicolumn{1}{c}{Acc.} 
    \\ \midrule
    GPT-4o-mini &66.61 &70.06 &69.61 &71.81 &69.52 
    \\
    \rowcolor{tablewhite} \textbf{w Triple} &\textbf{71.65} &\textbf{73.06} &\textbf{74.02} &\textbf{74.80} &\textbf{73.38 }
     \\
    GPT-4  &68.35 &70.54 &72.28 &72.28 &70.86 
    \\
    \rowcolor{tablewhite} \textbf{w Triple} &\textbf{71.71} &\textbf{73.41} &\textbf{75.89} &\textbf{75.45 }&\textbf{74.12} 
     \\
    \hline\hline
    \addlinespace[0.5ex]
    
    DeepSeek-R1 &68.35 &70.91 &72.76 &71.97 &70.74 
     \\
    \rowcolor{tablewhite} \textbf{w Triple} &\textbf{72.43} &\textbf{73.67} &\textbf{76.54} &\textbf{75.12} &\textbf{74.44} 
     \\
    Qwen2.5-72B-Ins.&61.94 &63.51 &66.05 &66.37 &64.47 
     \\
    \rowcolor{tablewhite} \textbf{w Triple} &\textbf{62.58} &\textbf{63.04} &\textbf{65.73 }&\textbf{66.21} &\textbf{64.39 }
    \\ 
    Llama3.3-70B-Ins. &61.94 &62.73 &64.48 &65.26 &63.60 
    \\
    \rowcolor{tablewhite} \textbf{w Triple} &\textbf{61.79} &\textbf{62.73} &\textbf{64.48} &\textbf{65.10} &\textbf{63.53 }
     \\ 
    \hline\hline
    \addlinespace[0.5ex]
    
    DS-R1-Dist-Qwen-32B &58.65 &60.58 &62.35 &63.17 &61.19 \\
    \rowcolor{tablewhite} \textbf{w Triple} &\textbf{62.17} &\textbf{63.25} &\textbf{65.82} &\textbf{66.04 }&\textbf{64.32} \\
    Qwen3-32B & 57.87 &60.36 &59.91 &62.44 &60.15 \\ 
    \rowcolor{tablewhite} \textbf{w Triple} &\textbf{61.39 }&\textbf{61.89} &\textbf{64.28} &\textbf{65.25} &\textbf{63.21}  \\ 
    Qwen2.5-32B-Ins. &58.82 &60.22 &61.02 &61.97 &60.51 \\
    \rowcolor{tablewhite} \textbf{w Triple} &\textbf{60.96 }&\textbf{63.33} &\textbf{65.15} &\textbf{66.37} &\textbf{63.44}  \\ 
    
    InternLM2.5-20B-Chat &54.41 &56.91 &59.61 &59.92 &57.71 \\
    \rowcolor{tablewhite} \textbf{w Triple}&\textbf{56.78} &\textbf{59.16} &\textbf{63.37 }&\textbf{61.04} &\textbf{59.53}   \\

    Qwen3-14B &56.57 &58.54 &59.81 &60.37 &58.04  \\ 
    \rowcolor{tablewhite} \textbf{w Triple} &\textbf{61.28 }&\textbf{61.48} &\textbf{66.01} &\textbf{65.07 }&\textbf{63.46 } \\

    Qwen2.5-14B &57.40 &59.44 &61.47 &60.06 &59.64  \\ 
    \rowcolor{tablewhite} \textbf{w Triple} &\textbf{60.17} &\textbf{60.30} &\textbf{64.55} &\textbf{64.02} &\textbf{62.18}  \\ 
    \hline\hline
    \addlinespace[0.5ex]
    
    Qwen3-8B  &54.88 &56.59 &59.61 &58.50 &57.40  \\ 
    \rowcolor{tablewhite} \textbf{w Triple} &\textbf{59.89} &\textbf{61.77 }&\textbf{64.00} &\textbf{63.63} &\textbf{62.07}  \\ 
    Qwen2.5-7B-Ins. &57.20 &57.07 &58.75 &58.40 &57.87  \\ 
    \rowcolor{tablewhite} \textbf{w  Triple} &\textbf{58.89} &\textbf{59.70} &\textbf{65.15} &\textbf{63.38} &\textbf{61.47}  \\ 
    
    DS-R1-Dist-Qwen-14B &57.15 &59.82 &60.76 &61.25 &59.75 \\
    \rowcolor{tablewhite} \textbf{w Triple} &\textbf{61.04} &\textbf{61.23} &\textbf{65.48} &\textbf{64.86} &\textbf{63.15} \\
    
    InternLM3-8B-Instruct  &\textbf{53.15} & \textbf{55.96} &58.03 &59.61 &56.69 
    \\ 
    \rowcolor{tablewhite} \textbf{w Triple} &51.25 &53.53 &\textbf{61.57} &\textbf{62.72} &\textbf{57.29} 
     \\ 
    InternLM2.5-7B-Chat  &53.32 &55.75 &65.18 &62.95 &59.98 \\ 
    \rowcolor{tablewhite} \textbf{w Triple} &\textbf{55.32} &\textbf{57.75} &\textbf{67.18} &\textbf{64.95} &\textbf{61.98 }\\ 
    
    \bottomrule
    \end{tabular}
    \end{threeparttable}
    \caption{Multichoice QA accuracy scores of LLMs. The input to LLMs is the current story plots. w / Triple indicates the prompt includes the character's ToM-based relation triples. Best performance of each model is bolded
    }
    \label{tab:auto_eva}
    \end{table*}

\begin{table*}[htb]
    \footnotesize
    \centering
    \begin{threeparttable}[b]
    \setlength{\tabcolsep}{15pt}  
    \begin{tabular}{lccccc}
    \toprule
    \multirow{2}{*}{\textbf{Models}}
    & \textbf{Belief} & \textbf{Desire} & \textbf{Emotion} & \multicolumn{1}{c}{\textbf{Intention}} & \textbf{Avg} \\
    &\multicolumn{1}{c}{Acc.} & \multicolumn{1}{c}{Acc.} & \multicolumn{1}{c}{Acc.} & \multicolumn{1}{c}{Acc.} & \multicolumn{1}{c}{Acc.}
    \\ \midrule
    GPT-4o-mini &68.66 &70.69 &72.28 &72.59 &71.05 \\
    \rowcolor{tablewhite} \textbf{w Triple} &\textbf{71.50} & \textbf{73.64} & \textbf{75.54} &\textbf{75.85} & \textbf{74.13} \\
    GPT-4 &67.87 &71.64 &74.17 & \textbf{75.75} &72.36
    \\
    \rowcolor{tablewhite} \textbf{w Triple} & \textbf{70.87} & \textbf{72.53} & \textbf{75.54} &75.22 & \textbf{73.54}
    \\
    \hline\hline
    \addlinespace[0.5ex]
    
    DeepSeek-R1 &68.76 &70.22 &72.49 &72.43 &70.98
    \\
    \rowcolor{tablewhite} \textbf{w Triple} &\textbf{71.81} &\textbf{73.85} &\textbf{75.85} &75.01 &\textbf{74.13} \\
    Qwen2.5-72B-Ins. &62.50 &63.61 &66.15 &65.99 &64.56
    \\
    \rowcolor{tablewhite} \textbf{w Triple} & \textbf{63.07} & \textbf{64.32} & \textbf{67.01} & \textbf{66.85} & \textbf{65.31}
    \\
    Llama3.3-70B-Ins. &61.47 &63.80 &65.77 &65.42 &64.12
    \\
    \rowcolor{tablewhite} \textbf{w Triple} & \textbf{62.76} & \textbf{64.27} & \textbf{67.17} & \textbf{66.69} & \textbf{65.22}
    \\
    \hline\hline
    \addlinespace[0.5ex]
    
    DS-R1-Dist-Qwen-32B &66.24 &68.35 &70.42 &71.19 &69.05 \\
    \rowcolor{tablewhite} \textbf{w Triple} & \textbf{70.56} & \textbf{72.43} & \textbf{74.85} & \textbf{74.97} & \textbf{73.20} \\
    Qwen3-32B & \textbf{61.72} & \textbf{63.05} & \textbf{66.51} & 66.37 & \textbf{64.41} \\
    \rowcolor{tablewhite} \textbf{w Triple} &60.91 &62.12 &66.36 & \textbf{67.21} &64.15 \\
    Qwen2.5-32B &61.81 &64.79 &66.85 & \textbf{67.01} &65.12 \\
    \rowcolor{tablewhite} \textbf{w Triple} & \textbf{62.13} & \textbf{64.95} & \textbf{66.69} &66.85 & \textbf{65.16} \\
    InternLM2.5-20B-Chat &56.73 &58.87 &63.84 &62.89 &60.58
    \\
    \rowcolor{tablewhite} \textbf{w Triple} &\textbf{58.30} & \textbf{60.42} & \textbf{64.44} & \textbf{63.32} & \textbf{61.62}
    \\
    
    Qwen3-14B &52.03 &53.41 &56.28 &56.32 &54.51 \\
    \rowcolor{tablewhite} \textbf{w Triple} &\textbf{54.05} &\textbf{55.10} &\textbf{58.29} &\textbf{58.33} &\textbf{56.44} \\
    Qwen2.5-14B-Ins. &51.81 &52.11 &57.17 &57.17 &54.57 \\
    
    \rowcolor{tablewhite} \textbf{w Triple} &\textbf{53.81} &\textbf{53.80} &\textbf{59.17} &\textbf{58.69} &\textbf{56.37} \\
    \hline\hline
    \addlinespace[0.5ex]
    
    Qwen3-8B &49.22 &51.76 &54.94 &55.09 &52.75 \\
    \rowcolor{tablewhite} \textbf{w Triple} &\textbf{51.82} & \textbf{54.79} & \textbf{58.12} & \textbf{58.28} & \textbf{55.75} \\
    Qwen2.5-7B-Ins. &51.34 &52.90 &56.54 &54.80 &53.90 \\
    \rowcolor{tablewhite} \textbf{w Triple} &\textbf{54.02} & \textbf{55.74} & \textbf{59.54} & \textbf{58.28} & \textbf{56.90} \\
    DS-R1-Dist-Qwen-14B &53.26 &54.89 &58.15 &58.68 &56.25 \\
    \rowcolor{tablewhite} \textbf{w Triple} &\textbf{57.85} & \textbf{59.47} & \textbf{63.26} & \textbf{63.75} & \textbf{61.08} \\
    InternLM3-8B-Ins. &50.35 &51.95 &55.19 &55.36 &53.21
    \\
    \rowcolor{tablewhite} \textbf{w Triple}&\textbf{54.87} &\textbf{55.60} &\textbf{59.31} &\textbf{59.72} &\textbf{57.38}
    \\
    InternLM2.5-7B-Chat &50.35 &51.95 &55.19 &55.36 &53.21
    \\
    \rowcolor{tablewhite} \textbf{w Triple} &\textbf{54.87} &\textbf{55.60} &\textbf{59.31} &\textbf{59.72} &\textbf{57.38}
    \\
    \bottomrule
    \end{tabular}
    \end{threeparttable}
    \caption{Multichoice QA performances of LLMs in terms of accuracy. The input to LLMs is the current story plots and previous plots' summary. Best performance of each model is bolded.
    }
    \label{tab:auto_eva_prev}
    \end{table*}

\begin{table}[htb]
    \footnotesize
    \centering
    \begin{threeparttable}[b]
    \resizebox{0.99\linewidth}{!}{
    \begin{tabular}{lccccc}
    \toprule
    \multirow{2}{*}{\textbf{Models}}
    & \textbf{Belief} & \textbf{Desire} & \textbf{Emotion} & \multicolumn{1}{c}{\textbf{Intention}} & \textbf{Avg} \\
    &\multicolumn{1}{c}{Acc.} & \multicolumn{1}{c}{Acc.} & \multicolumn{1}{c}{Acc.} & \multicolumn{1}{c}{Acc.} & \multicolumn{1}{c}{Acc.}
    \\ \midrule
    \multicolumn{6}{l}{{\cellcolor[rgb]{0.933, 0.933, 0.933}}\textbf{Direct Inference}} \\
    Qwen3-8B &51.10 &50.58 &53.31 &53.83 &52.21 \\
    \textbf{w Triple} &50.77 &50.72 &55.38 &55.90 &53.20  \\
    
    Qwen2.5-7B-Ins.  &53.85 &52.27 &57.40 &53.33 &54.21  \\
    \textbf{w Triple} &54.36 &52.27 &57.44 &53.39 &54.34\\
    
    InternLM3-8B-Ins. &50.40 &48.64 &54.35 &52.66 &51.51 \\
    \textbf{w Triple}&50.81 &50.76 &54.59 &52.97 &52.29 
  \\
    
    \multicolumn{6}{l}{{\cellcolor[rgb]{0.933, 0.933, 0.933}}\textbf{Fine-Tuning}} \\
    Qwen3-8B &53.22 &53.76 &54.94 &55.09 &54.25 \\
    \textbf{w Triple} &\textbf{59.57} &\textbf{57.50} &\textbf{58.74 }&\textbf{56.67} &\textbf{58.12} 
 \\
    Qwen2.5-7B-Ins. &55.22 &56.29 &56.82 &56.93 &56.32 
 \\
    \textbf{w Triple} &\textbf{59.91} &\textbf{57.73} &\textbf{58.12} &56.93 &\textbf{58.17}
\\
    InternLM3-8B-Ins. &55.40 &56.64 &57.35 &57.26 &56.64 
\\\
    \textbf{w Triple} &\textbf{58.91} &\textbf{58.73} &\textbf{58.12} &\textbf{58.93 }&\textbf{58.67} 
\\

\bottomrule
\end{tabular}
}
\end{threeparttable}
\caption{Ablation study results on out-of-distribution testsets across four ToM dimensions. "w Triple" indicates models that use structured triple representation in either inference or training.
} 
\label{tab:ablation}
\vspace{-1em}
\end{table}

\subsection{Out-of-Distribution Evaluation}

To evaluate the generalizability of \ourmethod~to new literary works, we conducted experiments using five books as an out-of-distribution (OOD) test set, comprising 779 questions across the four ToM dimensions. This setup allowed us to assess how well models augmented with \ourmethod~'s structured representations can transfer their ToM reasoning capabilities to entirely new narrative contexts not seen during training or development.
For these experiments, we selected three representative smaller-scale models: Qwen3-8B, Qwen2.5-7B-Instruct, and InternLM3-8B-Instruct. We evaluated each model in two distinct settings:

\noindent\textbf{Direct Inference.} Models were provided with the story plot, conversation scenario description, and question without any fine-tuning. We tested both standard inference (using only narrative content) and \ourmethod~-enhanced inference (including relevant mental state triples in the context).

\noindent\textbf{\ourmethodtitle-based Fine-Tuning.} Models were fine-tuned on training data where the output format first presented the relevant character relation triples followed by the correct answer option. This structured approach was designed to help models learn the explicit connections between narrative information, character mental states, and appropriate answers.
The \ourmethod~-based fine-tuning approach offers a significant advantage: it guides models to first extract structured knowledge representations before generating answers, effectively decomposing the complex ToM reasoning process into more manageable steps. By learning to generate structured triples as an intermediate step, models develop a more robust understanding of character psychology that transfers more effectively to new literary contexts.
Results from these experiments are presented in Table~\ref{tab:ablation}, demonstrating how the \ourmethod~-based approaches impact performance across different model architectures when faced with previously unseen literary works. We provide the training examples in Appendix~\ref{app:dataset}.

\section{Results and Analysis}
\subsection{Performance on ToM Reasoning Tasks}

The experimental results demonstrate the significant impact of \ourmethod~'s structured mental state triples across various ToM reasoning dimensions. As shown in Table~\ref{tab:auto_eva}, the integration of triple representations consistently enhances model performance, with improvements observed across all model scales and ToM dimensions. With an average prompt length of 2,500 tokens for both standard and \ourmethod~-enhanced inputs, these improvements highlight the value of structured representation rather than simply increasing context length.

The \ourmethod-enhanced approach yields substantial performance gains for all evaluated models. DeepSeek-R1 shows the most dramatic improvement, increasing from 70.74\% to 74.44\% when incorporating \ourmethod~triples. Similarly, Qwen3-14B experiences a remarkable improvement of 5.42\%, from 58.04\% to 63.46\%. Even top-performing models like GPT-4o benefit from \ourmethod~integration, improving from 70.86\% to 73.36\%. These consistent enhancements highlight the fundamental value of \ourmethod~'s structured knowledge representations in ToM reasoning tasks.

The impact of \ourmethod~is particularly pronounced for emotion recognition, where models show the largest accuracy gains. InternLM2.5-7B-Chat improves by 2.00\% in emotion accuracy, from 65.18\% to 67.18\%, while Qwen3-14B sees a remarkable improvement of 6.20\%, from 59.81\% to 66.01\%. This suggests that \ourmethod~'s explicit structured representations effectively bridge the gap between textual cues and the abstract emotional states they signify.
Notably, \ourmethod~integration partially mitigates the performance gap between smaller and larger models. While Qwen3-32B outperforms Qwen3-8B by 2.75\% in standard settings, this gap narrows when both incorporate \ourmethod~triples. This demonstrates how \ourmethod~'s structured knowledge representations can enhance the reasoning capabilities of smaller models, making sophisticated ToM reasoning more accessible.
\ourmethod~integration also helps balance performance across different ToM dimensions. Without triples, models typically perform best on Intention and worst on Belief, with considerable performance disparities. \ourmethod~integration narrows these gaps, providing more consistent reasoning capabilities across all mental state dimensions. For instance, DeepSeek-R1's performance spread between its strongest and weakest dimensions decreases from 4.41\% to 4.11\% with \ourmethod~enhancement.

\subsection{Performance with Extended Context}

Table~\ref{tab:auto_eva_prev} presents model performance when the input is expanded to include both current story plots and summaries of previous plots, increasing the average prompt length to approximately 4,500 tokens. This extended context scenario reveals important insights about model behavior with longer narratives and the continued effectiveness of \ourmethod~integration under more challenging conditions.
The addition of previous plot summaries creates a more challenging reasoning environment for all models, with notable performance decreases compared to the current-plot-only scenario in Table~\ref{tab:auto_eva}. For example, Qwen3-14B's accuracy drops substantially from 58.04\% to 54.51\%, and Qwen3-8B declines from 57.40\% to 52.75\%. This performance degradation reflects the well-known challenge LLMs face with longer contexts, where relevant information must be identified within a larger text span.
The integration of \ourmethod~'s structured mental state triples provides substantial benefits in this more challenging extended context scenario. DS-R1-Dist-Qwen-14B shows a dramatic improvement from 56.25\% to 61.08\%, while InternLM3-8B-Instruct improves from 53.21\% to 57.38\%. This demonstrates the robust utility of \ourmethod~'s structured representations in guiding model attention toward relevant character information across longer narrative spans.
The benefits of \ourmethod~integration are particularly evident for smaller models, which typically struggle more with extended contexts. Models like Qwen2.5-7B-Instruct show substantial improvements with triples, suggesting that \ourmethod~'s explicit structured knowledge helps these models overcome their inherent limitations in handling longer texts.
Performance patterns across ToM dimensions remain consistent with the current-plot-only scenario, with Emotion and Intention dimensions yielding higher accuracy than Belief and Desire dimensions. \ourmethod~integration helps narrow these dimensional performance gaps, providing more balanced reasoning capabilities.

\subsection{Ablation Study}

To assess the generalizability of \ourmethod~, we conducted an ablation study using five books as out-of-distribution test cases. These books were not part of the training data, allowing us to evaluate how well models transfer ToM reasoning capabilities to entirely new literary contexts.
As shown in Table~\ref{tab:ablation}, we compare two inference strategies across three model architectures. In the Direct Inference setting, models show modest performance on ToM reasoning tasks, with \ourmethod~-enhanced inference consistently outperforming standard inference across all dimensions. This confirms that \ourmethod~'s structured triple representation provides effective scaffolding for ToM reasoning even without task-specific training.
The Fine-Tuning section demonstrates significantly stronger results, where models were trained on data consisting of questions, \ourmethod~'s structured mental state triples, and answers. This triple-based training approach yields substantial improvements across all models and dimensions. For example, Qwen3-8B improves from 54.25\% to 58.12\% average accuracy when fine-tuned with \ourmethod~triples, and InternLM3-8B-Instruct shows the most dramatic improvement, reaching 58.67\% average accuracy.
The consistent performance gains across different architectures highlight the transferability of \ourmethod~to novel literary works. Notably, \ourmethod~fine-tuned models maintain balanced performance across all four ToM dimensions, suggesting that the triple-based representation effectively bridges the gap between different types of mental state reasoning.

\section{Conclusion}

We present \ourmethod~, a structured knowledge representation framework for enhancing Theory-of-Mind reasoning in narrative comprehension. Our character-centric ToM benchmark and perspective-aware temporal knowledge graph transform implicit character psychology into explicit relation triples that evolve throughout narratives. Experiments demonstrate that \ourmethod~significantly enhances reasoning capabilities across model scales and in extended-context scenarios, particularly helping smaller models bridge performance gaps with larger ones.

\section*{Ethical Statement}
Our benchmark uses literary works from the public domain Gutenberg Project, ensuring proper attribution and copyright compliance. The selected texts span different historical periods and cultural contexts, providing diverse examples of character psychology. Human annotators participating in the verification process were fairly compensated according to standard rates and fully informed about the task nature. We implemented a two-stage verification process to mitigate individual biases in interpretation.
We recognise that computational approaches to character understanding inevitably encode particular cultural perspectives or interpretive biases. Literary interpretation varies across cultural traditions, and our framework may reflect Western conceptions of psychology more prominently. While our research aims to advance fundamental capabilities in narrative comprehension, we acknowledge the broader implications for artificial systems that can model human mental states, emphasizing the importance of developing such technologies within frameworks that prioritize transparency and responsible use.

\section*{Limitations}
Our approach presents several limitations. First, reliance on GPT-4o for triple extraction introduces potential biases in character psychological profiles, as the model may favor certain interpretations over others or miss subtle contextual cues present in the original text. Second, our focus on four ToM dimensions (belief, emotion, intention, desire) doesn't capture other important aspects such as recursive beliefs (beliefs about others' beliefs), counterfactual reasoning, or epistemic states like uncertainty. Third, the structured triple format necessarily simplifies the complex, ambiguous nature of literary character psychology—for instance, a character's conflicted emotions or unconscious motivations may not fit neatly into subject-predicate-object structures. Finally, our multiple-choice evaluation, while allowing for systematic assessment, restricts measurement to recognition rather than testing deeper generative understanding of character psychology.

\bibliography{custom, zotero}

\appendix
\clearpage
\setcounter{table}{0}
\renewcommand{\thetable}{A\arabic{table}}
\setcounter{figure}{0}
\renewcommand{\thefigure}{A\arabic{figure}}

\section{Dataset Statistical}
\label{sec:dataset_statistics}

\subsection{Book Selection and Characteristics}

We selected 20 books from the CoSER dataset for the construction of our LitCharToM benchmark. These books from the Gutenberg Project are publicly accessible and span different historical periods, literary styles, and genres. Table~\ref{tab:book_statistics} lists the chosen books along with their plot counts, conversation counts, and average character numbers. Our benchmark features a diverse collection of 258 plots containing 599 conversations across these works. Notably, these books encompass a wide range of characters crafted by different authors with varying literary traditions. These characters possess distinct personalities, motivations, and backgrounds, representing diverse psychological profiles from ambitious royalty to contemplative philosophers. This diversity helps mitigate potential biases related to literary style, historical period, and cultural perspective while ensuring comprehensive coverage of different ToM reasoning challenges across narrative contexts.
The statistics for books we selected in this paper are shown in \autoref{tab:book_statistics} and \autoref{tab:ood_test_statistics}. \textbf{Detailed statistics of \ourdataset is shown in Table~\ref{tab:dataset_stats}}

\begin{table*}[ht]
\begin{center}
\begin{tabular}{lccc} 
\hline
 Book Name & Plots Num & Conversations Num & Avg Character \\ 
 \midrule
 King Lear & 14 & 42 & 3.00 \\
 A Study in Scarlet (Sherlock Holmes, \#1) & 14 & 41 & 2.73 \\
 The Scarlet Letter & 11 & 37 & 3.36 \\
 The Taming of the Shrew & 10 & 29 & 2.90 \\
 The Merchant of Venice & 11 & 33 & 3.00 \\
 The Tempest & 7 & 23 & 3.29 \\
 Julius Caesar & 7 & 20 & 2.86 \\
 The Call of the Wild & 8 & 22 & 2.75 \\
 A Portrait of the Artist as a Young Man & 12 & 30 & 2.50 \\
 The Wind in the Willows & 14 & 37 & 2.64 \\
 A Little Princess & 14 & 31 & 2.21 \\
 The Importance of Being Earnest & 14 & 36 & 2.57 \\
 Othello & 9 & 26 & 2.36 \\
 Dr Jekyll and Mr Hyde & 9 & 20 & 2.00 \\
 The Hound of the Baskervilles & 15 & 47 & 2.61 \\
 Notes from Underground & 19 & 37 & 1.85 \\
 The Turn of the Screw & 20 & 42 & 2.10 \\
 Jude the Obscure & 24 & 48 & 2.00 \\
 Siddhartha & 15 & 30 & 2.00 \\
 Anthem & 11 & 18 & 1.64 \\
 Total & 258 & 599 & 2.47 \\
\bottomrule
\end{tabular}
\caption{Statistics for the 20 books used in the evaluation.}
\label{tab:book_statistics}
\end{center}
\end{table*}

\begin{table*}[ht]
\begin{center}
\begin{tabular}{lccc} 
\hline
 Book Name & Plots Num & Conversations Num & Avg Character \\ 
 \midrule
 The Hound of the Baskervilles & 15 & 47 & 2.61 \\
 Notes from Underground & 19 & 37 & 1.85 \\
 The Turn of the Screw & 20 & 42 & 2.10 \\
 Jude the Obscure & 24 & 48 & 2.00 \\
 Siddhartha & 15 & 30 & 2.00 \\
 Total & 93 & 204 & 2.19 \\
\hline
\end{tabular}
\caption{Statistics for the 5 books used as out-of-distribution test set.}
\label{tab:ood_test_statistics}
\end{center}
\end{table*}

\subsection{Dataset Quality Control}
\label{app:quality}

To ensure data quality, we conduct a rigorous two-stage verification process for both questions and character relation triples. For the ToM-based questions, GPT-4o first verifies all generated questions for logical consistency, clarity, and the presence of a single unambiguously correct answer. Subsequently, human annotators assess a substantial portion of the questions for accuracy, difficulty level, and appropriateness, achieving a verification accuracy of 92.47\%. For the triple extraction, we employ a similar two-stage approach, with GPT-4o conducting an initial assessment followed by human expert verification of 40\% randomly selected triples, resulting in 93.64\% accuracy. Questions or triples identified as problematic during either verification stage undergo refinement or complete regeneration, followed by an additional verification cycle. This iterative process ensures the reliability and correctness of our benchmark for evaluating ToM reasoning capabilities in literary contexts.

\subsection{LitCharToM Dataset Statistics}

Our LitCharToM benchmark comprises a diverse collection of literary content for evaluating ToM reasoning capabilities. The dataset includes 20 books spanning different literary periods and genres, with 2,539 multiple-choice questions focused on character psychology. Each question is accompanied by one correct answer and three plausible distractor options, resulting in a total of 10,156 answer choices (2,539 correct answers and 7,617 distractors).

\begin{table}[ht]
\centering
\resizebox{0.99\linewidth}{!}{
\begin{tabular}{lcc}
\hline
\textbf{Dataset Characteristics} & \textbf{Count/Value} \\
\hline
Books & 20 \\
Questions & 2,539 \\
Correct Answers & 2,539 \\
Distractor Answers & 7,617 \\
\hline
\end{tabular}
}
\caption{Core statistics of the LitCharToM dataset.}
\label{tab:dataset_stats}
\end{table}

We evaluate models in two context settings: standard and extended. In the standard setting (current plot only), the average context length is 2,109 tokens, with a median of 2,094 tokens. For the extended setting (including previous plot summaries), the average context length increases substantially to 4,524 tokens, with contexts ranging from 1,259 to 20,366 tokens. This range of context lengths allows us to systematically evaluate how models handle ToM reasoning across different narrative scopes.

\begin{table}[ht]
\centering
\resizebox{0.99\linewidth}{!}{
\begin{tabular}{lcc}
\hline
\textbf{Context Length} & \textbf{Standard Setting} & \textbf{Extended Setting} \\
\hline
Average & 2,109 & 4,524 \\
Median & 2,094 & 2,894 \\
Minimum & 1,734 & 1,259 \\
Maximum & 2,601 & 20,366 \\
\hline
\end{tabular}
}
\caption{Context length statistics across different evaluation settings.}
\label{tab:context_stats}
\end{table}

\begin{figure}[ht]
\centering
\includegraphics[width=0.99\linewidth]{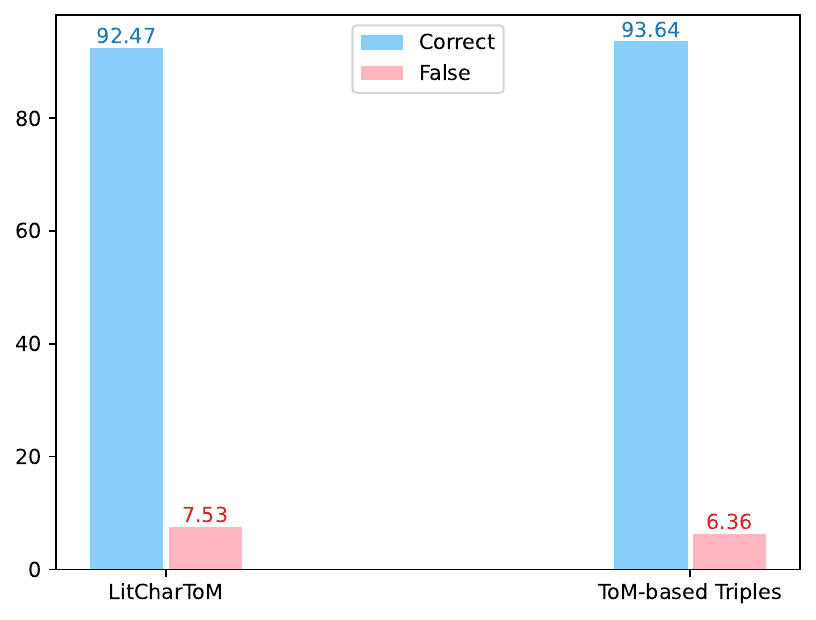}
\caption{Evaluation of generated data quality for LitCharToM dataset and ToM-based triples. Correct refers to the data verified as accurate by human annotators.}
\label{fig:data_quality}
\end{figure}

\section{Prompts}
\label{app:prompt}
\subsection{Prompt for Multiple Choice Question Generation}
\label{prompt:question}
The prompt for ToM-based multiple choice question generation is shown in Table~\ref{tab:question}.
\begin{table*}[b]
    \footnotesize
    \centering
\begin{tabular}{@{}p{\linewidth}@{}}
\toprule

\textbf{Prompt for Multiple Choice Question Generation} \\
\midrule
You are an expert in narrative analysis and character psychology, specializing in the application of Theory of Mind (ToM).\\
\\
Your task is to generate one multiple choice question for each of the following ToM dimensions — Belief, Emotion, Intention, and Desire — based on the provided story plot, scenario, character dialogues, and previous character relation triples. Each question must probe the psychological state of the Target Character, supported by reasoning grounded in both previously identified mental state triples and the current context.\\
\\
\# Definitions of Theory of Mind Dimensions:\\
\\
$<$Belief$>$: What the character believes to be true — this includes both objective facts and their subjective perceptions.\\
\\
$<$Emotion$>$: What the character feels — their affective responses, including joy, anger, fear, etc.\\
\\
$<$Intention$>$: What the character plans or wants to do — immediate or long-term actions driven by goals or motivations.\\
\\
$<$Desire$>$: What the character yearns for or wants to obtain — internal wishes, cravings, or goals (emotional or material).\\
\\
\# Input Fields:\\
\\
Plot summary: Contextual background of the narrative.\\
Current Scenario: The specific scene or moment in focus.\\
Dialogues: The words spoken and actions taken by characters in the scene.\\
Target Character: The character whose mental states are being analyzed.\\
Previous Character Relation Triples: Previously established mental state triples for the target character.\\
\\
\# Output Instructions:\\
1. For each ToM dimension, select relevant mental state triples.\\
\\
2. Construct one complex multiple choice question that requires reasoning and inference, not surface recall.\\
\\
3. Provide four answer options:\\
   - One correct answer, grounded in the character's psychology.\\
   - Three plausible but incorrect distractors, based on common misreadings or partial understanding.\\
\\
4. Do not repeat the same idea across different options.\\
\\
Output Format:\\
\{\\
\ "Target Character": [\\
\ \ \{"Belief Multiple Choice Question": \{\\
\ \ \ "Scenario": "xxx", "Reasoning":"xxx", "Question": "xxx",\\
\ \ \ "Options": ["A.xxx", "B.xxx", "C.xxx", "D.xxx"],\\
\ \ \ "Correct Answer": "x"\}\},\\
\ \ \{"Emotion Multiple Choice Question": \{...\}\},\\
\ \ \{"Intention Multiple Choice Question": \{...\}\},\\
\ \ \{"Desire Multiple Choice Question": \{...\}\}\\
\ ]\\
\}\\
\bottomrule
\end{tabular}
    \caption{Prompt for Multiple Choice Question Generation.}
    \label{tab:question}
\end{table*}

\subsection{Prompt for Character Relation Triple Generation}
\label{prompt:triple}
\begin{table*}[t]
\footnotesize
\centering
\begin{tabular}{p{0.98\linewidth}}
\hline
\multicolumn{1}{l}{\textbf{Prompt for Character Relation Triple Generation}} \\
\hline
You are an expert in analyzing narrative texts and understanding character psychology through the lens of Theory of Mind. 
Your task is to extract the triples of beliefs, emotions, intentions, and desires of a specific target character from the provided story plot summary, current scenario, and dialogues. 
You will output each identified mental state as a subject-predicate-object triple.
\\
Here are the definitions of the Theory of Mind dimensions you should use:
\\
\# ToM dimensions:
<Belief>: Beliefs encompass both objective facts and subjective perceptions concerning the existence or truth of something. 
\\
<Emotion>: Emotions are strong feelings deriving from one’s circumstances, mood, or relationships with others. And emotions are  variously associated with thoughts, feelings, behavioral responses, and a degree of pleasure or displeasure. 
\\
<Intention>: Intentions are blueprints that steer actions, encompassing both future plans and the motivations driving current behaviour. 
\\
<Desire>: Desires encompass both physical needs and psychological yearnings. Desires incline people toward action and fulfilling desires is pleasurable. Their fulfillment is normally experienced as pleasurable in contrast  to the negative experience of failing to do so.
\\
Analyze the provided Dialogues, the Target Character's explicitly stated Thoughts (if available in square brackets), and their Actions (if available in parentheses) within the context of the Story Plot Summary and Current Scenario.
\\
Identify instances of the Target Character's Beliefs, Emotions, Intentions, and Desires based on the definitions provided above.
\\
Output each identified mental state as a triple in the format: (Target Character, Predicate, Object).
Predicate should clearly indicate the ToM dimension (e.g., Believes, Feels, Intends, Desires) and can include a brief description of the target of the mental state (e.g., Believes about Cordelia's silence).
Object should be the content of the mental state (e.g., Cordelia's silence is a sign of disrespect and rebellion).
\\
This predicate can be further specified to provide more context, for example, using "BelievesAbout" to indicate a belief concerning another entity or event, or "FeelsTowards" to denote an emotion directed at someone or something. 
\\
Prioritize information that is directly attributable to the Target Character through their explicitly stated thoughts, actions, or spoken words.
\\
\# Example 
\\\\
<Plot summary>\\
In King Lear's palace, Kent and Gloucester discuss the King's preference between Albany and Cornwall. Lear, deciding to divide his kingdom among his daughters, $...$ 
\\
<Current Scenario>\\
In the opulent grand hall of King Lear's palace, anticipation hangs thick in the air.  $...$ 
\\
\# Dialogues between characters:\\
Environment: King Lear's grand hall, with courtiers and family gathered, as Lear prepares to speak."
\\
"King Lear: [I must know which daughter loves me most.] Tell me, my daughters, which of you shall we say doth love us most?
\\
"Goneril: Sir, I love you more than words can wield the matter; dearer than eyesight, space and liberty. Cordelia: (remains silent)"
\\
"King Lear: [She speaks well.] Of all these bounds, we make thee lady. What says our second daughter, Regan?"
\\
"Regan: I am made of that self metal as my sister, and prize me at her worth.Cordelia: Then poor Cordelia! And yet not so; since I am sure my love’s more ponderous than my tongue. $...$ 
\\
\# Target Character: King Lear 
\\
\# Output:
\\
\{\{
    "Target Character":
        [
            (King Lear, DesiresToKnow, which daughter loves King Lear most),
            (King Lear, IntendsTo, divide the kingdom based on his daughters' declarations of love),
            (King Lear, BelievesAboutCordelia, Cordelia's silence is a sign of defiance and disrespect),
            (King Lear, FeelsTowardsCordelia, wounded and betrayed by Cordelia's refusal to flatter King Lear),
            (King Lear, BelievesAboutGoneril, Goneril speaks well and expresses her love convincingly),
            (King Lear, FeelsTowardsCordelia, disappointed and shocked by Cordelia's honesty)
        ]
\}\}

Based on the provided Story Plot Summary, Current Scenario, and Dialogues, identify all relevant beliefs, emotions, intentions, and desires of the Target Character
Do not use pronoun in Object, use the name of Target character instead of his/her/them
Output each identified mental state as a triple in the format like (Target Character, Predicate, Object) in the following format.
\\\\
\# Input 
\\
<Plot summary>
\\
<Current Scenario>
\\
\# Dialogues between characters:
\\
\# Target Character: 
\\
\# Previous Character Triples:
\\
When analyzing the current scenario, consider the character's previously identified mental states triples from earlier plots. 
Your task is to:
\\\\
1. Integrate previous triples with your current analysis
2. For similar predicates (e.g., multiple beliefs about the same subject), combine or refine them based on new information
3. For conflicting predicates, update with the current information to reflect character development
4. Maintain consistency in the character's psychological profile while acknowledging changes in their mental states
\\
Use double quotes for all keys and values in the JSON. Do NOT include any explanation, markdown formatting, or additional comments Only return the JSON object.
You MUST return the result strictly in JSON format:

\# Output:
\{\{
    "Target Character":
            [
            "(Target Character, Predicate, Object)",
            "(Target Character, Predicate, Object)",
            "(Target Character, Predicate, Object)",
            "(Target Character, Predicate, Object)"
            ]
\}\} \\

\bottomrule
\end{tabular}
\caption{Prompt for Character Relation Triple Generation.}
\label{tab:triple}
\end{table*}

The prompt for ToM-based character relation triple generation is shown in Table~\ref{tab:triple}.

\begin{table*}[t]
    \centering
    \resizebox{0.89\linewidth}{!}{%
    \begin{tabular}{p{0.33\linewidth}p{0.33\linewidth}p{0.33\linewidth}}
    \toprule
    \multicolumn{3}{l}{\textbf{OOD Evaluation Input and Gold Triples}} \\
    \midrule
    \multicolumn{3}{p{\linewidth}}{%
You are an expert in narrative analysis and character psychology, specializing in Theory of Mind (ToM). Your task is to analyze the mental states of characters in literary works.
\medskip

For the character "Siddhartha" in the book "Siddhartha", analyze their mental state based on the following context:

\medskip

STORY PLOT:

Siddhartha, a Brahmin's son, grows up with his friend Govinda. He excels in spiritual practices and is loved by all. However, he becomes dissatisfied with traditional teachings and seeks a deeper understanding of the self and the universe.

\medskip
SCENARIO:

In the serene landscape of ancient India, beneath the sprawling branches of a majestic banyan tree, Siddhartha and Govinda sit in a tranquil meditation space. The early morning light filters through the leaves, casting soft shadows on the ground. The surrounding area is peaceful, with distant sounds of chirping birds and a gentle breeze rustling the leaves. The setting represents a pivotal moment in their spiritual journey, where traditional teachings are about to be challenged by Siddhartha's profound inner questioning.

\medskip
QUESTION:

What is Siddhartha's intention regarding his spiritual journey?

CANDIDATE CHOICES:

A. To continue following traditional teachings without question.

B. To seek a deeper understanding of Atman beyond words and rituals.

C. To convince Govinda to abandon their current path together.

D. To establish himself as a new spiritual leader among the Brahmins.%

\medskip
      (Siddhartha, DesiresToExpress, doubts to Govinda without shaking Govinda's faith)
      
      (Siddhartha, Feels, weight of questions growing during meditation)
      
      (Siddhartha, BelievesAboutTeachings, traditional teachings are profound but lacking)
      
      (Siddhartha, Desires, to truly know Atman beyond words and rituals)
      
      (Siddhartha, FeelsTowardsGovinda, touched by Govinda's loyalty but worried for Govinda)
      
      (Siddhartha, IntendsToSeek, further understanding beyond traditional teachings)
      
      (Siddhartha, BelievesAboutPath, true enlightenment might lie beyond current teachings)
      
      (Siddhartha, DesiresToFindTruth, whatever the cost)    } \\
    \midrule
    \multicolumn{3}{l}{\textbf{What is Siddhartha's intention regarding his spiritual journey?}} \\
    \midrule
    \textbf{InternLM3-8B-Ins.} &
    \textbf{InternLM3-8B-Ins. (w triples)} &
    \textbf{Qwen3-8B} \\[3pt]
    
    Relevant mental state triples:
    
    \medskip
    (Siddhartha, IntendsTo, express his inner doubts and desire for 
    deeper understanding to Govinda)
    
    \medskip
    (Siddhartha, IntendsTo, seek the ultimate truth beyond words and 
    rituals)
    
    \medskip
    Answer:\{answer: C\} &

    \medskip
    Relevant mental state triples:

    \medskip
    (Siddhartha, IntendsTo, explore the nature of the self beyond traditional teachings)

    \medskip
    Answer:\{answer\: B\} &
   
    Relevant mental state:
    
    \medskip
    (Siddhartha, IntendsTo, seek a deeper understanding of Atman and the universe)

    \medskip
    (Siddhartha, IntendsTo, set aside dogue's teachings and embrace the path of light together with Govinda)

    \medskip
    Answer:\{answer: C\}
    \\
    \midrule
    \textbf{Qwen3-8B (w triples)} &
    \textbf{Qwen2.5-7B-Ins.} &
    \textbf{Qwen2.5-7B-Ins. (w triples)} \\[3pt]
    Relevant mental state:

    \medskip
    (Siddhartha, IntendsTo, understand the true self beyond traditional teachings)

    \medskip
    (Siddhartha, IntendsTo, approach the spiritual journey with an open mind)

    \medskip
    Answer:\{answer: B\}
    &
    Relevant mental state triples:
    
    \medskip
    (Siddhartha, IntendsTo, question the nature of Atman and the purpose of life)

    \medskip
    (Siddhartha, IntendsTo, embark on a new path of self-discovery and spiritual exploration)

    \medskip
    Answer:\{answer: A\}
    &
    Relevant mental state triples:
    
    \medskip
    (Siddhartha, IntendsTo, question the teachings of the oldest Brahmin)
    \medskip
    (Siddhartha, IntendsTo, pursue the understanding of Atman beyond words and rituals)
    
    \medskip
    Answer:\{answer: B\}\\
    \bottomrule
    \end{tabular}%
    }
    \caption{Model predictions for book Siddhartha's intention question of OOD evaluation.}
    \label{tab:case-study-1-1}
\end{table*}

\section{Dataset Examples}
\label{app:dataset}

\subsection{OOD Evaluation Results}

Table~\ref{tab:case-study-1-1} presents detailed model predictions for a representative question from our OOD test set, demonstrating how \ourmethod~'s structured triples influence model reasoning. When comparing models with and without triple information, we observe that triple-enhanced models consistently identify Siddhartha's deeper spiritual intentions more accurately. While InternLM3-8B generates the correct answer even without triples, Qwen3-8B and Qwen2.5-7B-Ins only arrive at the correct answer when provided with explicit triple representations. This pattern illustrates how \ourmethod~'s structured knowledge helps bridge reasoning gaps, particularly for complex questions requiring nuanced understanding of character motivations across extended narrative contexts.

\subsection{Training Set}
The training examples for two different experiment setting for OOD evaluation are shown
in  Table~\ref{tab:training_example1} to  Table~\ref{tab:training_example2}.

\begin{table*}[htbp]
    \scriptsize
    \centering
\begin{tabular}{@{}p{\linewidth}@{}}
\toprule
\textbf{Training example for OOD evaluation} \\
\midrule
\#\# Input: \\\\
You are an expert in narrative analysis and character psychology, specializing in Theory of Mind (ToM). Your task is to analyze the mental states of characters in literary works.
\\\\
For the character King Lear in the book King Lear, analyze their mental state based on the following context:
\\\\
STORY PLOT:\\
King Lear decides to divide his kingdom among his three daughters based on their professions of love. Cordelia, refusing to flatter, is disinherited. Kent is banished for defending her. Lear gives his power to Goneril and Regan, who soon begin to undermine his authority. The Fool criticizes Lear's actions, and Lear starts to realize his mistake. Goneril demands Lear reduce his retinue, leading to a heated confrontation. Lear decides to leave for Regan's castle.
\\\\
SCENARIO:\\
The royal court is assembled in a grand hall, filled with tension and anticipation. Ornate tapestries and gilded decorations surround the throne where King Lear sits, preparing to divide his kingdom. Goneril and Regan have already delivered their exaggerated declarations of love, receiving large portions of the kingdom. Now, all eyes are on Cordelia, the youngest daughter, as she prepares to speak. The atmosphere is charged with expectation, silence hanging heavy as courtiers watch the pivotal moment that will determine the future of the kingdom.
\\\\
QUESTION:\\
What does King Lear believe about Cordelia's profession of love?
\\\\
CANDIDATE CHOICES:\\
A. He believes she is jesting and will eventually flatter him.\\
B. He believes she is being honest and true to herself.\\
C. He believes she is intentionally defying him out of spite.\\
D. He believes she is confused and doesn't understand the situation.\\
\\\\
First, identify the relevant mental state triples (beliefs, emotions, intentions, or desires) that explain King Lear's psychology in this scenario.\\
Then, based on these mental states, select the most appropriate answer from the choices above.
\\\\
Format your response as:\\
1. List the relevant mental state triples\\
2. Provide your answer as a JSON object: \{answer: X\} where X is the letter (A, B, C, or D) of the correct choice.    
\\\\
\#\# Output:\\\\
"Relevant mental state triples:\\
(Siddhartha, IntendsTo, express his inner doubts and desire for deeper understanding to Govinda)\\
(Siddhartha, IntendsTo, seek the ultimate truth beyond words and rituals)\\
Answer:\\
\{answer: B\}

\\

\bottomrule
\end{tabular}
    \caption{Example of training data with triples.}
    \label{tab:training_example1}
\end{table*}

\begin{table*}[htbp]
    \scriptsize
    \centering
\begin{tabular}{@{}p{\linewidth}@{}}
\toprule
\textbf{Training example for OOD evaluation} \\
\midrule
\#\# Input: \\\\
You are an expert in narrative analysis and character psychology, specializing in Theory of Mind (ToM). Your task is to analyze the mental states of characters in literary works.
\\\\
For the character King Lear in the book King Lear, analyze their mental state based on the following context:
\\\\
STORY PLOT:\\
King Lear decides to divide his kingdom among his three daughters based on their professions of love. Cordelia, refusing to flatter, is disinherited. Kent is banished for defending her. Lear gives his power to Goneril and Regan, who soon begin to undermine his authority. The Fool criticizes Lear's actions, and Lear starts to realize his mistake. Goneril demands Lear reduce his retinue, leading to a heated confrontation. Lear decides to leave for Regan's castle.
\\\\
SCENARIO:\\
The royal court is assembled in a grand hall, filled with tension and anticipation. Ornate tapestries and gilded decorations surround the throne where King Lear sits, preparing to divide his kingdom. Goneril and Regan have already delivered their exaggerated declarations of love, receiving large portions of the kingdom. Now, all eyes are on Cordelia, the youngest daughter, as she prepares to speak. The atmosphere is charged with expectation, silence hanging heavy as courtiers watch the pivotal moment that will determine the future of the kingdom.
\\\\
QUESTION:\\
What does King Lear believe about Cordelia's profession of love?
\\\\
CANDIDATE CHOICES:\\
A. He believes she is jesting and will eventually flatter him.\\
B. He believes she is being honest and true to herself.\\
C. He believes she is intentionally defying him out of spite.\\
D. He believes she is confused and doesn't understand the situation.\\
\\\\
First, identify the relevant mental state triples (beliefs, emotions, intentions, or desires) that explain King Lear's psychology in this scenario.\\
Then, based on these mental states, select the most appropriate answer from the choices above.
\\\\
Format your response as:\\
1. List the relevant mental state triples\\
2. Provide your answer as a JSON object: \{answer: X\} where X is the letter (A, B, C, or D) of the correct choice.    
\\\\
\#\# Output:\\\\
Answer:\\
\{answer: B\}

\\

\bottomrule
\end{tabular}
    \caption{Example of training data w/o triples.}
    \label{tab:training_example2}
\end{table*}

\end{document}